\PassOptionsToPackage{colorlinks=true,linkcolor=blue,citecolor=blue,urlcolor=blue}{hyperref}
\documentclass[a4paper,fleqn]{cas-dc}
\usepackage[authoryear]{natbib}

\makeatletter
\@ifundefined{figref}{\newcommand{\figref}[1]{\hyperref[#1]{Fig.~\ref*{#1}}}}{\renewcommand{\figref}[1]{\hyperref[#1]{Fig.~\ref*{#1}}}}
\@ifundefined{Figref}{\newcommand{\Figref}[1]{\hyperref[#1]{Figure~\ref*{#1}}}}{\renewcommand{\Figref}[1]{\hyperref[#1]{Figure~\ref*{#1}}}}
\@ifundefined{tabref}{\newcommand{\tabref}[1]{\hyperref[#1]{Table~\ref*{#1}}}}{\renewcommand{\tabref}[1]{\hyperref[#1]{Table~\ref*{#1}}}}
\@ifundefined{Tabref}{\newcommand{\Tabref}[1]{\hyperref[#1]{Table~\ref*{#1}}}}{\renewcommand{\Tabref}[1]{\hyperref[#1]{Table~\ref*{#1}}}}
\@ifundefined{figreftwo}{\newcommand{\figreftwo}[2]{\hyperref[#1]{Figs.~\ref*{#1}} and \hyperref[#2]{\ref*{#2}}}}{\renewcommand{\figreftwo}[2]{\hyperref[#1]{Figs.~\ref*{#1}} and \hyperref[#2]{\ref*{#2}}}}
\@ifundefined{tabreftwo}{\newcommand{\tabreftwo}[2]{\hyperref[#1]{Tables~\ref*{#1}} and \hyperref[#2]{\ref*{#2}}}}{\renewcommand{\tabreftwo}[2]{\hyperref[#1]{Tables~\ref*{#1}} and \hyperref[#2]{\ref*{#2}}}}
\newcommand{\subfigref}[2]{\hyperref[#1]{Fig.~\ref*{#1}(#2)}}
\makeatother

\ExplSyntaxOn
\RenewDocumentEnvironment { Abstract } { o }
 {
  \group_begin:
  \IfNoValueTF{#1}{}{ \tex_gdef:D \abstractname {#1} }
  \parindent=0pt\relax
  \box_if_empty:NTF \g_stm_key_box
   { \leftskip=.35\textwidth\relax }
   {
    \dim_set:Nn \l_tmpa_dim { \box_ht:N \g_stm_key_box + \box_dp:N \g_stm_key_box }
    \leftskip=.35\textwidth\relax
    \hspace*{-.35\textwidth}
    \makebox[0pt][l]{\box_use_drop:N \g_stm_key_box}
    \skip_vertical:n { -\l_tmpa_dim }
   }
  \noindent \abstractname \par
  \skip_vertical:n { -4pt }
  \noindent \rule{.65\textwidth}{.2pt}\par\footnotesize
  \ignorespaces \everypar { \parindent=1.5em\relax }
 }
 { \par \group_end: }
\ExplSyntaxOff

\ExplSyntaxOn
\cs_set:Npn \__make_fig_caption:nn #1#2
{
  \skip_vertical:N \l_fig_abovecap_skip
  \setbox\cascaptionbox=\hbox{%
     \rmfamily\small\textbf{Fig.\thefigure{}:}~#2}
  \ifdim\the\wd\cascaptionbox<\dim_use:N \l_fig_width_dim\relax
    \parbox{\l_fig_width_dim}
      {\unskip\ignorespaces\hfil\rmfamily\small
       \textbf{Fig.\thefigure{}:}~#2\hfil\par}
  \else
    \parbox{\l_fig_width_dim}
      {\rightskip=0pt\unskip\ignorespaces\rmfamily\small
       \textbf{Fig.\thefigure{}:}~#2\par}
  \fi
  \skip_vertical:N \l_fig_belowcap_skip
}
\ExplSyntaxOff

\begin{document}
\let\WriteBookmarks\relax
\def\floatpagepagefraction{1}
\def\textpagefraction{.001}
\let\printorcid\relax 

\shorttitle{MSR for Cervical CT--MRI Registration}

\shortauthors{Bohai Zhang et al.}

\title[mode=title]{MSR: Hybrid Field Modeling for CT--MRI Rigid--Deformable Registration of the Cervical Spine with an Annotated Dataset}

\author[1,2,3]{Bohai Zhang}
\ead{ssc.1230609@gmail.com}
\author[1,2,3]{Wenjie Chen}
\ead{smu22520357@smu.edu.cn}
\author[4]{Mu Li}
\ead{271841418@qq.com}
\author[1,2,3]{Kaixing Long}
\ead{2016301626@qq.com}
\author[5]{Xing Shen}
\ead{sxfrancis@163.com}
\author[5]{Xinqiang Yao}
\ead{yxq123@smu.edu.cn}
\author[5]{Jincheng Yang}
\ead{gdgkyjc@126.com}
\author[5]{Jianting Chen}
\ead{chenjt@smu.edu.cn}
\author[1,2,3]{Wei Yang}
\ead{weiyanggm@gmail.com}
\author[1,2,3]{Qianjin Feng}
\ead{fengqj99@smu.edu.cn}
\author[1,2,3]{Lei Cao\cormark[1]}
\ead{caolei@smu.edu.cn}

\address[1]{School of Biomedical Engineering, Southern Medical University, Guangzhou, 510515, China.}
\address[2]{Guangdong Provincial Key Laboratory of Medical Image Processing, Guangzhou, 510515, China.}
\address[3]{Guangdong Province Engineering Laboratory for Medical Imaging and Diagnostic Technology Guangzhou, 510515, China.}
\address[4]{Information Center, Nanfang Hospital, Southern Medical
University, Guangzhou, 510515, China.}
\address[5]{Division of Spine Surgery, Department of Orthopaedics, Nanfang hospital, Southern Medical University, Guangzhou, Guangdong, 510515, China.}

\begin{abstract}
Accurate CT--MRI registration of the cervical spine is essential for preoperative planning because this region is anatomically complex, highly variable, and vulnerable to injury of the vertebral arteries and spinal cord. However, cervical CT--MRI registration remains underexplored, particularly for rigid--deformable hybrid modeling, and the lack of high-quality annotated multimodal data further limits progress. To address these challenges, we construct and release a comprehensively annotated CT--MRI dataset, R-D-Reg, and propose MSR, a  rigid--deformable hybrid registration framework for complex joint structures. Specifically, MSR includes a rigid registration module for independent local rigid alignment of individual vertebrae and a deformable registration module with an MSL block that combines Mamba-based global modeling and Swin Transformer-based local modeling through adaptive gating. The rigid and deformable deformation fields are then fused to generate a hybrid field that better preserves local anatomical consistency. Experiments on the Neck, HN, and TH datasets show that MSR achieves the best mean Dice scores of 79.84\,$\pm$\,3.59\%, 96.43\,$\pm$\,1.13\%, and 88.30\,$\pm$\,4.62\%, respectively. The code and dataset are publicly available at \url{https://github.com/ssc1230609-spec/MSR-registration}.
\end{abstract}




\maketitle

\section{Introduction}

\begin{figure}
  \centering
  \includegraphics[width=\linewidth]{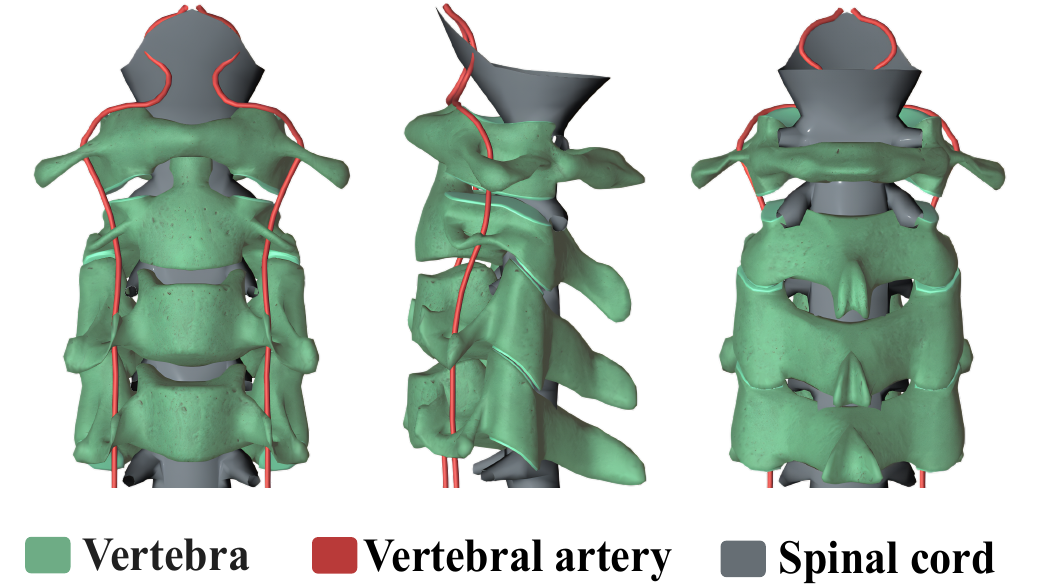}
  \caption{Adjacent vertebral arteries and spinal cord in the cervical spine region.From left to right: anterior, lateral, and posterior views.}
  \label{fig:cervical-anatomy}
\end{figure}

Spine CT--MRI registration is a representative multimodal registration task for articulated structures, with important applications in orthopedic image analysis \citep{garg2024bone}, preoperative planning \citep{rosenman1998image}, and image-guided interventions \citep{holly2006evaluation}. For example, atlantoaxial pedicle screw fixation is one of the most widely used internal fixation techniques in cervical spine surgery for relieving neural compression and restoring the stability of the atlantoaxial joint. However, the cervical spine is among the most anatomically complex and variable regions of the spine, comprising rigid vertebrae surrounded by soft tissues with substantial deformability, as illustrated in \figref{fig:cervical-anatomy}. As a result, surgical procedures in this region carry a considerable risk of inadvertent injury to adjacent critical structures, such as the vertebral arteries and spinal cord. In preoperative imaging, CT provides clear visualization of bony anatomy, whereas MRI offers superior soft-tissue contrast \citep{ryan2019pet}, making the two modalities highly complementary. Therefore, accurate registration of CT and MRI from the same patient is essential for precise surgical planning and risk reduction. Nevertheless, research on CT--MRI registration for the cervical spine remains limited, and high-quality annotated public datasets are still scarce, which hinders systematic method development and objective evaluation.

In medical image registration, existing methods are generally categorized as rigid or deformable according to the underlying spatial transformation model \citep{chen2025survey}. Rigid registration primarily models global translations and rotations. However, because of its limited degrees of freedom, conventional rigid or affine registration often fails to capture the local pose variations commonly observed in spinal joint regions. To address this limitation, piecewise rigid and hierarchical models \citep{little1997deformations,hu2004multirigid,rasoulian2010group} have been proposed to improve performance by modeling different anatomical structures independently. Nevertheless, these methods often rely on manually defined anatomical partitions, which limits their robustness and clinical applicability. In contrast, deformable registration \citep{balakrishnan2019voxelmorph} can represent complex nonlinear deformations and therefore provides greater flexibility. However, in spinal imaging scenarios, many deformable methods neglect the intrinsic rigidity of bony structures and simplify the problem as purely deformable modeling, leading to results that lack structural consistency and anatomical plausibility. Consequently, in practical spinal registration tasks, a single transformation model is insufficient to simultaneously satisfy the rigidity constraints of bones and the complex nonlinear deformations of soft tissues \citep{gao2024mairnet}. This challenge cannot be adequately resolved by global rigid registration alone, nor can it be fully captured by a purely deformable model; rather, it is inherently a rigid--deformable hybrid registration problem.

For rigid--deformable hybrid registration, existing studies typically handle the rigid component by introducing constraint terms into the loss function, leveraging segmentation masks for guidance, or designing multi-branch network architectures to inject rigid priors into deformable registration frameworks. However, these strategies still rely on the optimization process to approximate rigid transformations, making them sensitive to loss-weight balancing and the instability of cross-modal similarity measures. For deformable modeling, mainstream approaches are predominantly based on convolutional neural networks (CNNs) \citep{balakrishnan2019voxelmorph}. Although CNN-based methods are effective for learning local nonlinear deformations, they remain limited in capturing long-range dependencies among different anatomical structures in the cervical spine. In recent years, Transformer-based methods \citep{chen2022transmorph} that exploit self-attention have been introduced into deformable registration to enhance global spatial interactions. Meanwhile, state space models (SSMs), particularly the representative Mamba architecture \citep{gu2023mamba}, have also been explored for global dependency modeling because of their strength in sequence modeling. However, the potential of these two families of methods for rigid--deformable hybrid registration remains largely underexplored. In addition, the cervical spine exhibits substantial inter-subject variability, with complex and highly localized deformation patterns between bony structures and surrounding soft tissues, which further increases the difficulty of registration. Hierarchical vision Transformers such as Swin Transformer \citep{liu2021swin} employ window-based self-attention to enable efficient local feature interaction and therefore hold promise for modeling fine-grained structural variations. Nevertheless, their effectiveness in capturing localized cervical deformations still requires further validation. In summary, for cervical CT--MRI registration, how to jointly model the global rigidity of bony structures while achieving coordinated representation of global deformation trends and local structural alignment within a unified framework remains a key challenge for improving both registration accuracy and anatomical plausibility.

In this study, we propose a novel structure-aware rigid--deformable hybrid registration framework, termed MSR, to address the above challenges. The key idea is to explicitly model rigid and deformable transformations by generating a rigid deformation field through a dedicated rigid registration module and a deformable deformation field through a deformable registration module, followed by their fusion for unified rigid--deformable modeling. In addition, to meet the practical requirements of cervical CT--MRI registration, we construct a new benchmark dataset consisting of paired CT--MRI images. The main contributions of this work are summarized as follows:

\begin{itemize}
  \item We propose a mask-guided rigid deformation field generation method, in which the model learns the local rigid motions of bony structures to achieve independent alignment of each structure, thereby enabling rigid registration.
  \item We introduce an MSL module into the deformable registration stage. This module combines Mamba-based global modeling and Swin Transformer-based local modeling, and uses a gating mechanism for adaptive fusion to improve registration accuracy and anatomical plausibility.
  \item We construct and publicly release a CT--MRI multimodal registration dataset, R-D-Reg, containing paired CT/MRI data and structural annotations. The dataset is built using segmentation-derived masks with manual verification, thereby providing a unified and reproducible benchmark for evaluation.
  \item To the best of our knowledge, this is the first study to specifically address cervical CT--MRI registration within a rigid--deformable hybrid registration framework.
\end{itemize}

\section{Related work}

\subsection{Registration methods for spine}

For spine image registration, early studies recognized that neither global rigid registration nor purely deformable registration can simultaneously satisfy anatomical consistency and deformation expressiveness. This insight led to the development of rigid--deformable hybrid registration paradigms, in which the spine is modeled as a set of locally rigid units, while continuous deformation fields are used to capture the relative motion between these rigid structures and the elastic variations of the surrounding tissues.

\subsubsection{Traditional spine registration methods}

In traditional approaches, spine registration is typically formulated as a composition of multiple local rigid transformations. \citet{arsigny2006log} generated a global deformation field by fusing multiple local rigid transformations. \citet{luong2011incorporating} adopted a nonlinear programming-based strategy with hard constraints to estimate vertebra-wise rigid transformations, thereby enforcing strictly rigid motion in bony regions while allowing free deformation in non-rigid areas. \citet{vcech2006piecewise} proposed a two-stage segmentation-then-registration framework, in which anatomical priors were first used to estimate local rigid transformations, followed by smooth interpolation to construct the final deformation field. \citet{gill2012biomechanically} treated each vertebra as an independent rigid subvolume, allowing separate transformations rather than applying a single global transformation to the entire spine. Although these methods capture local rigidity to some extent, they generally rely on accurate segmentation or explicit geometric modeling and involve iterative optimization to couple rigid and deformable motion, resulting in limited computational efficiency. Moreover, in cross-modal scenarios, their robustness to initialization errors and large deformations remains limited.

\subsubsection{Learning-based spine registration methods}

In recent years, deep learning-based image registration methods have achieved substantial progress \citep{balakrishnan2019voxelmorph}. These approaches learn a direct mapping from image pairs to deformation fields, replacing traditional iterative optimization with global function approximation and therefore offering clear efficiency advantages during inference. In spinal image registration, several studies have attempted to incorporate rigid constraints into deep learning frameworks to improve anatomical plausibility. For example, \citet{jian2022weakly} embedded vertebra-wise local rigidity priors into a global deformable registration process through additional loss terms under weak supervision. \citet{zhao2023spineregnet} proposed an end-to-end framework, SpineRegNet, which uses CT and MRI segmentation masks to jointly estimate affine--elastic deformation fields within a unified network. \citet{gao2024mairnet} designed a dual-branch architecture, in which a non-learning multi-rigid branch provides anatomically consistent rigid priors, while a learnable deformable branch models continuous deformations, and the two branches are jointly optimized. \citet{tang2020admir} constrained a deformable registration network with an affine registration network and fused their outputs at the deformation-field level, enabling one-shot resampling in an end-to-end manner. Although these methods partly alleviate the lack of rigid constraints, they mainly rely on constraint injection or architectural design and rarely address the problem from the perspective of deformation-field generation. As a result, explicit modeling and coordinated fusion of local rigid motion and continuous deformation within a unified transformation space remain limited. Consequently, for complex spinal structures, existing methods still struggle to simultaneously ensure the geometric consistency of bony structures and the continuity of soft-tissue deformation. In addition, many of these methods rely on private datasets with segmentation annotations \citep{gao2024mairnet,zhao2023spineregnet}, whereas publicly available CT--MRI spine registration datasets remain scarce, especially those providing both high-quality registration pairs and fine-grained structural annotations. This limitation hinders reproducibility and fair evaluation.

Furthermore, the performance of learning-based methods strongly depends on feature representation. In CT--MRI cross-modal scenarios, the substantial differences in imaging mechanisms make it difficult for conventional convolutional models or any single modeling paradigm to capture both global structural consistency and local fine-grained variation, thereby limiting the accuracy and stability of deformation estimation. Therefore, developing a framework that jointly models global and local representations remains critical.

\subsection{State space models and Swin Transformer}

To enhance feature representation, recent studies have introduced novel architectures with complementary modeling capabilities. State space models (SSMs) \citep{gu2021efficiently} propagate information through recursive state updates, enabling the modeling of long-range contextual dependencies in sequential representations. This property underlies the application of Mamba in various image analysis tasks. For example, \citet{zhang2025convmamba} used Mamba to capture long-range dependencies from serialized features while employing convolutional networks to extract local spatial information. \citet{cheng2025mamba} adopted Mamba as the backbone of a segmentation network and designed global and local sequence enhancement strategies to improve feature representation. \citet{wang2025mamba} integrated Mamba into the encoder of a U-shaped network as the core sequence-modeling module for multimodal brain image registration. \citet{lin2025mamba} further constructed a parallel architecture in which a Mamba-based global branch and a convolutional branch jointly model global anatomical context. Despite their advantages in global structure modeling, SSMs rely on recursive information propagation that mainly emphasizes overall contextual relationships, which limits their ability to capture fine-grained local structures. As a result, they are less suitable for independently handling precise deformation estimation, particularly in regions with substantial local texture variation.

In contrast, the Swin Transformer uses a window-based attention mechanism to model pixel relationships within local windows, thereby enhancing the representation of local structural patterns. Its hierarchical shifted-window design and cross-window interaction strategy enable effective modeling of texture details and structural boundaries. \citet{iqbal2023bts} employed Swin Transformer as an auxiliary encoder to improve feature representation of complex structures via window-based attention. \citet{ye2025robust} further introduced this mechanism into medical image registration, leveraging a dual-branch Swin Transformer backbone to accurately capture local bone textures and structural boundaries in 2D--3D registration tasks.

Overall, SSMs excel at global structure modeling, whereas window-based self-attention is more effective for capturing fine-grained local representations. These two paradigms therefore exhibit strong complementarity in terms of modeling scope and information interaction. Recent studies have begun to explore their joint modeling. For example, \citet{ngu2025mstransbts} constructed a parallel architecture in MSTransBTS with a Mamba branch and a Swin Transformer branch, combining global SSM-based modeling and local window attention to extract multi-scale contextual features for accurate 3D brain tumor segmentation. \citet{liu2024mstfnet} adopted Mamba in MSTFNet to model global dependencies along the spectral dimension while employing a two-stage Swin Transformer to capture spatial local textures, achieving effective hyperspectral image classification through feature fusion. \citet{hatamizadeh2025mambavision} proposed MambaVision, a hierarchical architecture that leverages Mamba for efficient global modeling in early stages and introduces window-based self-attention blocks in later stages to enhance local detail representation and semantic integration. However, existing Mamba--Transformer hybrid approaches are primarily designed for visual recognition or segmentation tasks, and their fusion strategies mainly focus on parallel or cascaded interactions at the feature level. They are rarely tailored to deformation modeling in medical image registration. In cross-modal registration of complex articulated structures, it is necessary not only to preserve the global anatomical consistency of bony structures through global context modeling, but also to capture continuous and fine-grained local deformations of soft tissues. Therefore, how to effectively integrate global dependency modeling and local structural representation within a unified framework, and further map them to deformation-field estimation, remains a key challenge for improving feature representation and registration accuracy.

These observations motivate the design of a Mamba--Swin hybrid feature encoding strategy, in which global SSM-based features and local window-attention features are extracted in parallel and fused to enhance the consistency and discriminability of cross-modal representations. In addition, because the contributions of global and local features vary across anatomical regions, adaptive fusion between the two types of features is particularly important for articulated structures such as the cervical spine.

\begin{figure*}[t]
  \centering
  \includegraphics[width=0.9\textwidth]{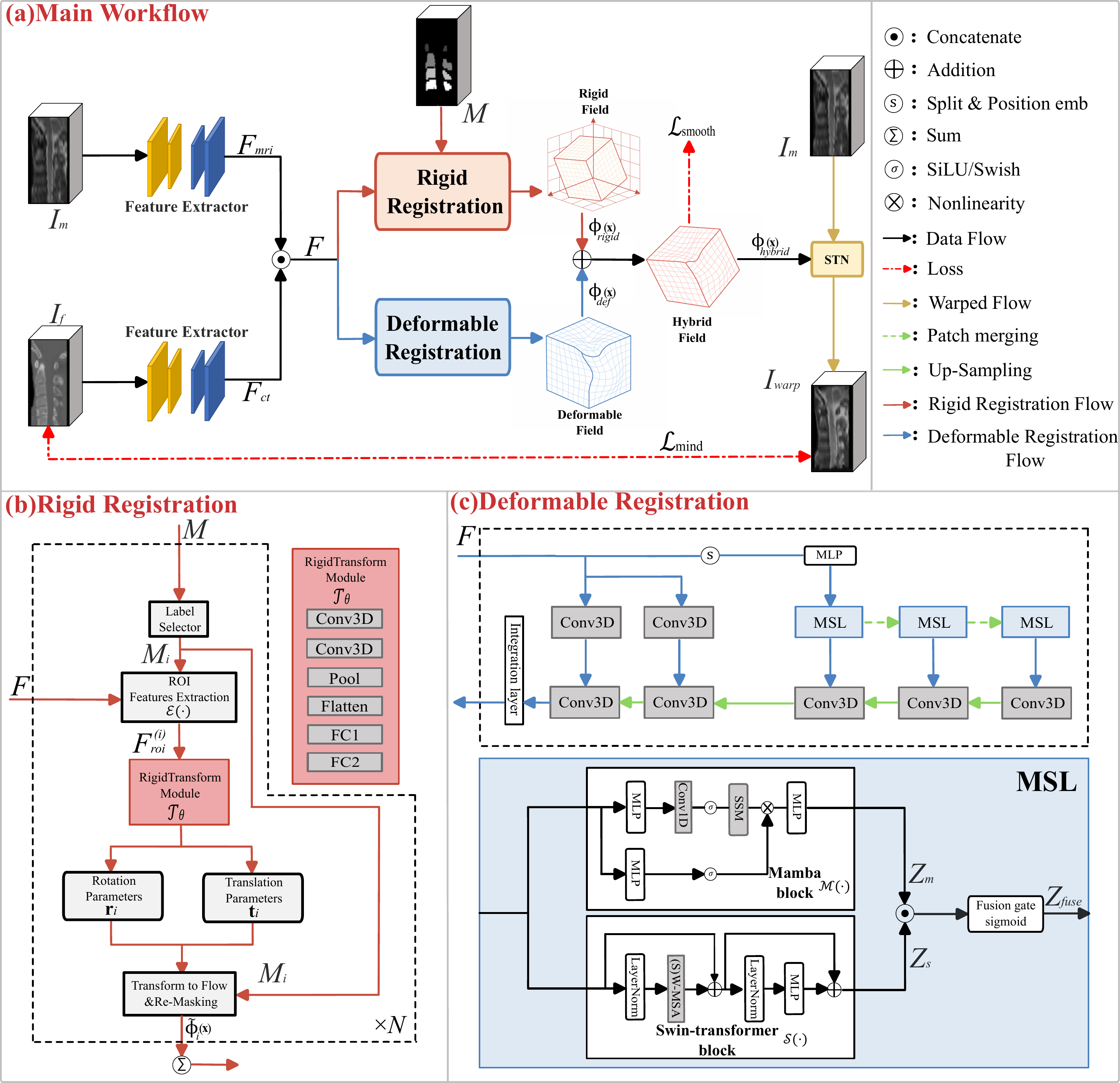}
  \caption{The overall framework of MSR.}
  \label{fig:msr-framework}
\end{figure*}

\section{Method}

\subsection{Overview}

We denote the MRI image as the moving image $I_m$ and the CT image as the fixed image $I_f$. The goal of registration is to learn a spatial transformation function $\Phi:\mathbb{R}^3 \rightarrow \mathbb{R}^3$ such that the transformed moving image $I_m \circ \Phi$ is structurally aligned with the fixed image $I_f$. Here, $\circ$ denotes the spatial resampling operation implemented via a Spatial Transformer Network (STN) \citep{jaderberg2015spatial}. As illustrated in \figref{fig:msr-framework}, the overall framework first extracts multimodal features $F_{\mathrm{ct}}$ and $F_{\mathrm{mri}}$, and then concatenates them along the channel dimension to obtain a unified feature representation $F$:
\begin{equation}
F = \mathrm{Concat}(F_{\mathrm{ct}}, F_{\mathrm{mri}})
\end{equation}

In cross-modal registration of the cervical spine, it is necessary to preserve the rigid consistency of bony structures while accurately modeling the continuous deformation of surrounding soft tissues. Therefore, we jointly consider rigid and deformable registration and construct a unified hybrid representation within a single framework. Specifically, the feature $F$ is fed into two parallel branches. On the one hand, a rigid registration module is employed to model the rigid motion of bony structures via rotation and translation, yielding a rigid deformation field. The rigid transformation at a spatial location $\mathbf{x} \in \mathbb{R}^3$ is defined as
\begin{equation}
\Phi_{\mathrm{rigid}}(\mathbf{x}) = \mathbf{R}\mathbf{x} + \mathbf{t}, \quad \mathbf{R} \in \mathrm{SO}(3),\; \mathbf{t} \in \mathbb{R}^3
\end{equation}
where $\mathbf{R}$ denotes the rotation matrix and $\mathbf{t}$ denotes the translation vector. On the other hand, a deformable registration module based on Mamba and Swin Transformer is adopted to capture both global and local structural variations. A dense displacement field $\mathbf{u}(\mathbf{x})$ is used to describe nonlinear spatial deformation, and the corresponding transformation is defined as
\begin{equation}
\Phi_{\mathrm{def}}(\mathbf{x}) = \mathbf{x} + \mathbf{u}(\mathbf{x})
\end{equation}
where $\mathbf{u}(\mathbf{x}): \mathbb{R}^3 \rightarrow \mathbb{R}^3$ represents the displacement vector at each spatial location. Finally, the rigid deformation field $\Phi_{\mathrm{rigid}}(\mathbf{x})$ and the deformable deformation field $\Phi_{\mathrm{def}}(\mathbf{x})$ predicted by the two branches are fused to obtain a unified hybrid transformation $\Phi_{\mathrm{hybrid}}$, which is applied to the moving image via STN:
\begin{equation}
I_{\mathrm{warp}} = I_m \circ \Phi_{\mathrm{hybrid}}
\end{equation}
Accordingly, the overall optimization objective can be formulated as finding the optimal $\Phi_{\mathrm{hybrid}}$ among all feasible transformations:
\begin{equation}
\Phi_{\mathrm{hybrid}}^{*} = \arg\min_{\Phi_{\mathrm{hybrid}}} \; \mathcal{L}(I_f, I_m \circ \Phi_{\mathrm{hybrid}})
\end{equation}
Where $\mathcal{L}$ is the unsupervised registration loss defined as
\begin{equation}
\mathcal{L} = \mathcal{L}_{\mathrm{sim}}(I_f, I_{\mathrm{warp}}) + \lambda \mathcal{L}_{\mathrm{smooth}}(\Phi_{\mathrm{hybrid}})
\end{equation}
Here, $\mathcal{L}_{\mathrm{sim}}$ denotes the image similarity measure (MIND), which enforces cross-modal structural consistency, and $\mathcal{L}_{\mathrm{smooth}}$ is a smoothness regularization term that constrains the spatial smoothness of the deformation field. 

In the following, we describe the two key components of the proposed framework: rigid registration and deformable registration.

\subsection{Rigid registration}

The cervical spine consists of multiple vertebrae, each exhibiting relatively independent rigid motion. To address the limitation of global rigid models in accurately capturing local structural motion, as shown in \subfigref{fig:msr-framework}{b}, we propose a mask-guided rigid registration module in the feature space. Each vertebra is assigned a label $i$, and rigid parameters are iteratively estimated for each vertebra. These parameters are then converted into local rigid displacement fields that are nonzero only within the corresponding vertebral region. Finally, all local fields are aggregated to form the overall rigid deformation field. The rigid registration module consists of the following four steps.

\noindent\textbf{Step 1: Label Selector.}
Let the segmentation labels be denoted as $M$, where each vertebra corresponds to a binary mask $M_i$ $(i=1,\ldots,N)$. A label selector is employed to sequentially extract the binary mask of each vertebra from the multi-label segmentation map. These masks are then mapped into the feature space and sequentially applied as $M_i$ to constrain the region of interest (ROI) for the current rigid estimation.

\noindent\textbf{Step 2: ROI Feature Extraction.}
Given the fused feature $F$ and mask $M_i$, mask-guided ROI feature extraction is performed to obtain
\begin{equation}
F_{\mathrm{roi}}^{(i)} = \mathcal{E}(F, M_i)
\end{equation}
where $\mathcal{E}(\cdot)$ denotes feature cropping and aggregation within the masked region. Specifically, ROI cropping is first performed using $M_i$, followed by global average pooling to aggregate local features into a compact representation, enabling the network to focus on vertebra-specific rigid motion.

\noindent\textbf{Step 3: RigidTransform Module.}
For each vertebra, the ROI feature $F_{\mathrm{roi}}^{(i)}$ is used to estimate rigid transformation parameters $T_\theta$. The module first employs two 3D convolutional layers to model local structural patterns and extract discriminative representations related to vertebral geometry. Then, adaptive global average pooling is applied to compress spatial features into a global descriptor, enhancing the modeling of overall rigid motion. Finally, a multilayer perceptron predicts the rotation and translation parameters:
\begin{equation}
(\mathbf{r}_i, \mathbf{t}_i) = T_\theta(F_{\mathrm{roi}}^{(i)}), \quad \mathbf{r}_i \in \mathbb{R}^3,\; \mathbf{t}_i \in \mathbb{R}^3
\end{equation}
where $\mathbf{r}_i$ and $\mathbf{t}_i$ denote the rotation and translation parameters, respectively.

\noindent\textbf{Step 4: Transform to Flow \& Re-masking.}
The rigid transformation of vertebra $i$ is defined as
\begin{equation}
\Phi_{\mathrm{rigid}}^{(i)}(\mathbf{x}) = \mathbf{r}_i \mathbf{x} + \mathbf{t}_i
\end{equation}
and the corresponding displacement field is
\begin{equation}
\phi_i(\mathbf{x}) = \Phi_{\mathrm{rigid}}^{(i)}(\mathbf{x}) - \mathbf{x}
\end{equation}
To ensure that rigid motion is only applied within the corresponding vertebral region, the displacement field is masked as
\begin{equation}
\tilde{\phi}_i(\mathbf{x}) = M_i \, \phi_i(\mathbf{x})
\end{equation}
Finally, all local displacement fields are summed to obtain the overall rigid deformation field:
\begin{equation}
\Phi_{\mathrm{rigid}}(\mathbf{x}) = \sum_{i=1}^{N} \tilde{\phi}_i(\mathbf{x})
\end{equation}
The resulting rigid deformation field is further fused with the deformable field to produce the final hybrid transformation.

\subsection{Deformable registration}

In the cervical spine region, complex nonlinear deformations arise from soft tissue motion, imaging discrepancies, and local structural mismatches, particularly at the interfaces between soft tissues and bone. To address this, we propose a deformable registration modeling strategy. As illustrated in \subfigref{fig:msr-framework}{c}, the deformable branch takes the fused feature $F$ as input. It first employs 3D convolutions to extract initial local feature representations and progressively constructs multi-scale features. During the encoding stage, features are successively processed by multiple stacked Mamba--Swin Layers (MSL), enabling progressive transformation and fusion, thereby enlarging the receptive field and enhancing feature representation capability. For each MSL, the input feature is denoted as $Z$, and the layer consists of two parallel branches:

\begin{equation}
Z_m = \mathcal{M}(Z)
\end{equation}

\begin{equation}
Z_s = \mathcal{S}(Z)
\end{equation}

where $\mathcal{M}(\cdot)$ denotes the Mamba block based on SSM, which models long-range dependencies via selective scanning over serialized features to capture global contextual information. $\mathcal{S}(\cdot)$ denotes the Swin Transformer block, which leverages window-based self-attention to model feature interactions within local windows and employs a shifted-window strategy to enable cross-window information exchange, thereby enhancing the representation of local structural patterns and boundary details.

On this basis, a gating mechanism is introduced to adaptively fuse the outputs $Z_m$ and $Z_s$. Specifically, the features are first concatenated along the channel dimension, followed by a linear projection and a Sigmoid activation to generate fusion weights:

\begin{equation}
G = \sigma\big(\mathcal{C}(\mathrm{Concat}(Z_m, Z_s))\big)
\end{equation}

where $\mathcal{C}(\cdot)$ denotes the linear mapping for gate generation, and $\sigma(\cdot)$ represents the Sigmoid function. The fused feature is then defined as:

\begin{equation}
Z_{\mathrm{fuse}} = G \odot Z_m + (1 - G) \odot Z_s
\end{equation}

where $\odot$ denotes element-wise multiplication. This gating mechanism enables the network to adaptively select between representations emphasizing global consistency or local structural sensitivity across different spatial regions. For instance, in regions with large-scale deformations, the model tends to rely more on the Mamba branch to maintain overall structural consistency; whereas in areas with complex local structures, such as boundaries, the Swin Transformer branch contributes more to improving local alignment accuracy. In this way, a balance between large-scale deformation modeling and fine-grained structural alignment is achieved. The fused feature $Z_{\mathrm{fuse}}$ is then passed to the next MSL, enabling progressive feature refinement across hierarchical layers.

After multiple MSL encoding stages, high-level fused feature representations are obtained. These features are then fed into the decoder, where spatial resolution is progressively recovered through upsampling and feature fusion, followed by 3D convolutional layers for feature reconstruction. Finally, a convolutional layer maps the features to a three-channel displacement field, yielding the deformable deformation field $\Phi_{\mathrm{def}}(\mathbf{x})$, which is further fused with the rigid deformation field to obtain the final hybrid transformation:

\begin{equation}
\Phi_{\mathrm{hybrid}}(\mathbf{x}) = \Phi_{\mathrm{def}}(\mathbf{x}) + \Phi_{\mathrm{rigid}}(\mathbf{x})
\end{equation}

\section{Material}

\subsection{Dataset construction}

This study is a retrospective study approved by the Institutional Review Board (IRB) and conducted in accordance with the Declaration of Helsinki, with the requirement for informed consent waived. The Neck dataset was collected from Guangdong Traditional Chinese Medicine Hospital and underwent strict de-identification, with all personally identifiable information removed.

To address the scarcity of publicly available data for cross-modal registration in complex articulated structures, we construct and release a CT--MRI registration dataset with structural annotations, termed R-D-Reg (Rigid--Deformable Registration). R-D-Reg is developed based on the dataset released in the SynthRAD 2025 Challenge by \citeauthor{thummerer2025synthrad2025} (\citeyear{thummerer2025synthrad2025}), which includes three anatomical regions: head and neck (HN), thorax (TH), and abdomen (AB). Owing to the difficulty of consistently selecting vertebral structures in the abdominal region, only the HN and TH subsets are included in the dataset construction.

In R-D-Reg, each case contains a paired CT volume, an MRI volume, and the corresponding anatomical segmentation masks. These annotations support registration methods that require segmentation guidance and facilitate the evaluation of rigid--deformable hybrid registration in terms of both accuracy and generalization. It should be noted that the publicly released dataset includes only the HN and TH subsets, whereas the Neck dataset from Guangdong Traditional Chinese Medicine Hospital is not publicly available.

\subsection{Data processing pipeline}

\begin{figure*}[t]
  \centering
  \includegraphics[width=0.9\textwidth]{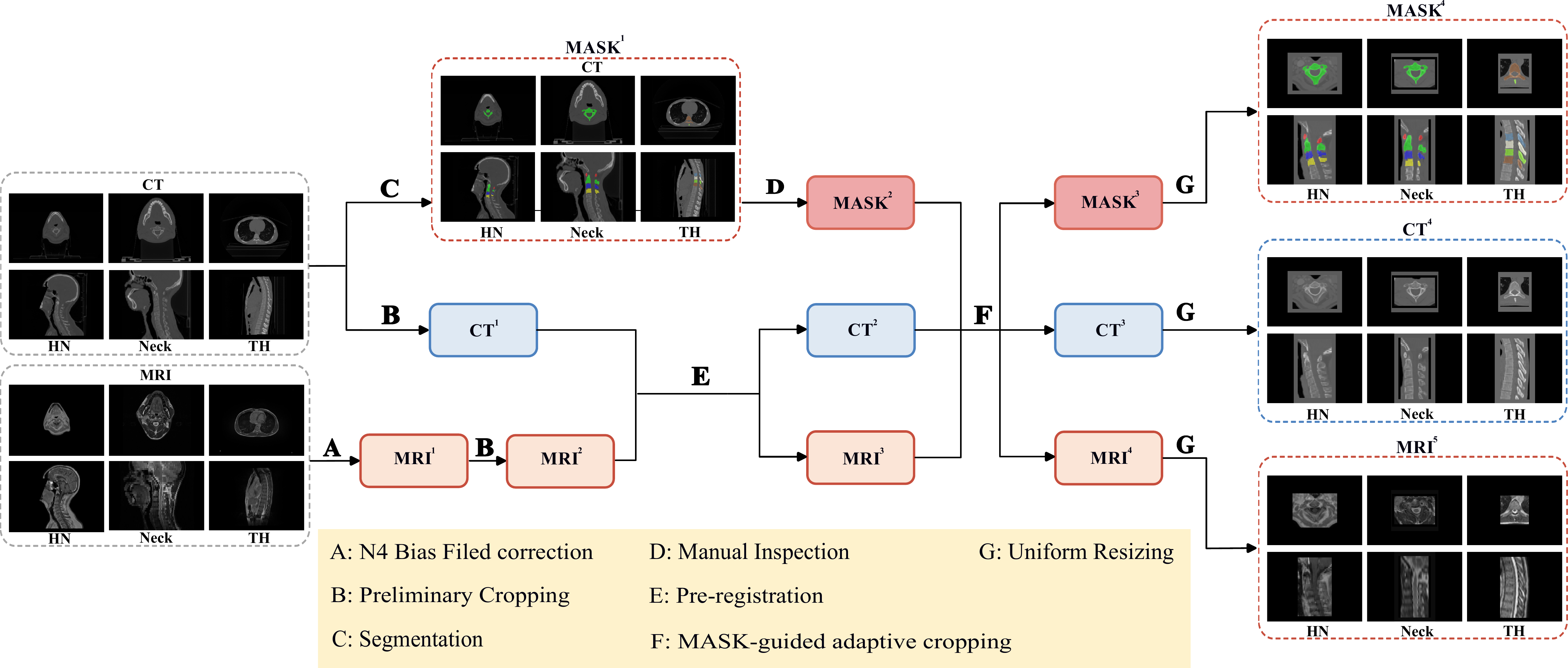}
  \caption{Data processing pipeline of the R-D-Reg dataset.}
  \label{fig:data-processing-pipeline}
\end{figure*}

As illustrated in \figref{fig:data-processing-pipeline}, the overall data processing pipeline consists of the following steps. (A) N4 bias field correction is applied to the raw MRI data to reduce intensity inhomogeneity caused by magnetic field non-uniformity and to improve cross-modal structural consistency. (B) The CT and MRI volumes are preliminarily cropped to remove large background regions and reduce the data extent, thereby improving computational efficiency for subsequent processing. (C) Anatomical structures are segmented to generate corresponding masks. Considering the relatively large mobility of vertebrae in the cervical region, nnU-Net \citep{isensee2021nnu} is employed for the HN and Guangdong Traditional Chinese Medicine Hospital (Neck) datasets. Specifically, a subset of cases is manually annotated to train the model, and once stable weights are obtained, the remaining data are automatically segmented to generate structural labels. For the TH dataset, TotalSegmentator \citep{wasserthal2023totalsegmentator} is used to automatically generate anatomical labels. (D) The segmentation masks are manually reviewed to ensure annotation quality. (E) An initial rigid pre-registration is performed between CT and MRI volumes to establish a unified spatial reference. (F) Based on the segmentation masks, anatomically relevant regions are extracted, and the CT and MRI volumes are cropped to retain valid regions, thereby defining the final field of view (FOV). (G) Finally, all volumes are resized to a unified spatial resolution to ensure consistent input for network training. The Neck, HN, and TH datasets are standardized to $[106,92,52]$, $[122,94,52]$, and $[106,128,74]$, respectively.

Through the above pipeline, we obtain the R-D-Reg CT--MRI registration dataset with complete structural annotations and consistent spatial alignment. The HN and TH subsets are publicly released, together with the corresponding segmentation model weights, to support the development and evaluation of cross-modal registration methods. Detailed dataset statistics are reported in \tabref{tab:dataset-statistics}. In addition, to enhance reproducibility and extensibility, the full data processing pipeline will also be made publicly available.

\begin{table}[t]
  \makeatletter
  \long\def\@makecaption#1#2{%
    \centering
    \vskip 6pt
    {\parbox{\columnwidth}{\rightskip=0pt\fontfamily{ptm}\selectfont\fontsize{10pt}{12pt}\selectfont\textbf{#1}\par#2\par\vskip4pt}}%
  }
  \makeatother
  \caption{Dataset statistics.}
  \label{tab:dataset-statistics}
  \centering
  {\fontfamily{ptm}\selectfont
  \fontsize{8pt}{12pt}\selectfont
  \renewcommand{\arraystretch}{1.2}
  \setlength{\tabcolsep}{2pt}
  \begin{tabular*}{\columnwidth}{@{\extracolsep{\fill}}>{\raggedright\arraybackslash}p{0.36\columnwidth}>{\centering\arraybackslash}p{0.18\columnwidth}>{\centering\arraybackslash}p{0.18\columnwidth}>{\centering\arraybackslash}p{0.18\columnwidth}@{}}
    \hline
    Dataset & Neck & HN & TH \\
    \hline
    Train/Validation/Test & 71/8/12 & 54/6/8 & 134/16/18 \\
    Size & [106,92,52] & [122,94,52] & [106,128,74] \\
    Spacing (mm$^3$) & $1 \times 1 \times 1$ & $1 \times 1 \times 3$ & $1 \times 1 \times 3$ \\
    Label & C1--C4 & C1--C4 & T6--T9 \\
    \hline
  \end{tabular*}
  \renewcommand{\arraystretch}{1}
  }
\end{table}

\section{Experiments and results}

\subsection{Implementation Details}

In this study, all models are trained for 500 epochs. The Adam optimizer is used to optimize the network parameters. The initial learning rate is set to $5 \times 10^{-5}$, the regularization weight $\lambda$ is set to 0.2, and the batch size is set to 2. All experiments are implemented in PyTorch 2.2.2 and conducted on an NVIDIA GeForce RTX 3090 GPU.

\subsection{Evaluation Metrics}

To quantitatively evaluate the registration performance of the proposed method, three commonly used metrics are adopted.

\noindent\textbf{Dice similarity coefficient (DSC)} \citep{pang2020spineparsenet}.
The Dice similarity coefficient is used to measure the overlap between annotated vertebrae before and after registration, thereby assessing the overall accuracy of the registration results. A higher DSC indicates better alignment between the transformed anatomical structures and the reference.

\noindent\textbf{Jacobian determinant ($J_{\phi}$)} \citep{rohlfing2003volume}.
To ensure the anatomical plausibility of the deformation, the Jacobian determinant is employed to evaluate the topological properties of the deformation field. Since folding is anatomically implausible, we compute the percentage of non-background voxels with negative Jacobian determinants. A lower percentage indicates smoother and more physically plausible deformations.

\noindent\textbf{95\% Hausdorff distance (HD95).}
The Hausdorff distance measures the boundary discrepancy between two sets. To more robustly assess boundary alignment, we adopt the 95\% Hausdorff distance (HD95), defined as the 95th percentile of distances between boundary points of the registered and reference anatomical structures. This metric is particularly sensitive to boundary errors and is important for evaluating alignment accuracy in regions such as soft tissue--bone interfaces. A lower HD95 indicates more accurate registration.

\subsection{Comparison Methods}

To evaluate the performance of the proposed method, we compare it with several state-of-the-art and classical registration approaches, including the following methods.

\noindent\textbf{Affine.}
Affine is an intensity-based affine registration method using mutual information (MI) as the similarity metric. A multi-resolution optimization strategy is adopted, with resolution levels set from 16 (coarsest) to 8 (finest).

\noindent\textbf{SyN.} \citep{avants2008symmetric}
SyN employs the Mattes mutual information metric with 32 histogram bins. The number of iterations is set to $100 \times 50 \times 20$, with Gaussian smoothing sigmas of $0.5 \times 0.2 \times 0$, and shrink factors of $4 \times 2 \times 1$.

\noindent\textbf{VoxelMorph.} \citep{balakrishnan2019voxelmorph}
VoxelMorph is a representative CNN-based deformable registration method. It adopts a U-Net architecture to learn the deformation vector field (DVF) in an end-to-end manner, modeling spatial correspondences via local convolution operations.

\noindent\textbf{TransMorph.} \citep{chen2022transmorph}
TransMorph replaces conventional CNN architectures with a Transformer--ConvNet hybrid framework, employing a Swin Transformer as the encoder to enhance feature representation and long-range interaction.

\noindent\textbf{MambaMorph.} \citep{wang2025mamba}
MambaMorph is a method based on state space models (SSMs), which leverages the Mamba architecture to model volumetric data as sequences, enabling effective modeling of complex spatial dependencies.

\noindent\textbf{TransMorph-rigid.}
Built upon the TransMorph framework, this variant incorporates the proposed rigid registration module and integrates it with the original deformable branch, forming an extended model with explicit rigid constraints. This method is used to analyze the effect of introducing explicit rigid modeling within a Transformer-based representation framework. It also serves as a baseline for comparison with the proposed MSL-based deformable module, validating the effectiveness of the rigid module under different deformable modeling strategies.

\noindent\textbf{MambaMorph-rigid.}
This variant integrates the proposed rigid registration module into the MambaMorph framework, enabling joint modeling with the original deformable process. It is used to evaluate the impact of incorporating rigid constraints within an SSM-based framework on registration accuracy and structural consistency. In addition, it serves as a comparative baseline against the proposed MSL-based deformable module, demonstrating the generality and effectiveness of the rigid module across different deformable modeling paradigms.

\subsection{Results and analysis}

\subsubsection{Comparative study on Neck and HN}

\begin{table*}[t]
  \makeatletter
  \long\def\@makecaption#1#2{%
    \centering
    \vskip 6pt
    {\parbox{\textwidth}{\rightskip=0pt\fontfamily{ptm}\selectfont\fontsize{10pt}{12pt}\selectfont\textbf{#1}\par#2\par\vskip4pt}}%
  }
  \makeatother
  \caption{Quantitative comparison on the Neck and HN test datasets. Results are reported as mean $\pm$ standard deviation (Mean $\pm$ Std). The best performance is highlighted in \textbf{bold}, and the second-best is \underline{underlined}. An upward arrow ($\uparrow$) indicates that higher values are better, while a downward arrow ($\downarrow$) indicates that lower values are better.}
  \label{tab:neck-hn-comparison}
  \centering
  {\fontfamily{ptm}\selectfont
  \fontsize{8pt}{9.8pt}\selectfont
  \renewcommand{\arraystretch}{1.18}
  \setlength{\tabcolsep}{3pt}
  \setlength{\extrarowheight}{1pt}
  \resizebox{\textwidth}{!}{%
  \begin{tabular*}{1.03\textwidth}{@{\extracolsep{\fill}}llccccccc@{}}
    \hline
    Datasets & Methods & \% of $|J| < 0$ $\downarrow$ & HD95 $\downarrow$ & C1 $\uparrow$ & C2 $\uparrow$ & C3 $\uparrow$ & C4 $\uparrow$ & Average Dice $\uparrow$ \\
    \hline
    \multirow{8}{*}{Neck} & Affine & -- & 2.45$\pm$1.44 & 0.646$\pm$0.092 & 0.672$\pm$0.113 & 0.589$\pm$0.223 & 0.540$\pm$0.253 & 61.17$\pm$15.61 \\
    & SyN \citep{avants2008symmetric} & 0.227$\pm$0.417 & 2.69$\pm$1.13 & 0.669$\pm$0.106 & 0.679$\pm$0.100 & 0.634$\pm$0.132 & 0.650$\pm$0.114 & 65.81$\pm$10.71 \\
    & VoxelMorph \citep{balakrishnan2019voxelmorph} & \textbf{<0.0001} & 2.35$\pm$0.95 & 0.669$\pm$0.131 & 0.676$\pm$0.136 & 0.659$\pm$0.169 & 0.705$\pm$0.154 & 67.71$\pm$14.13 \\
    & TransMorph \citep{chen2022transmorph} & \underline{0.079$\pm$0.097} & 1.63$\pm$0.39 & 0.763$\pm$0.066 & 0.791$\pm$0.065 & 0.771$\pm$0.064 & 0.775$\pm$0.052 & 77.53$\pm$5.33 \\
    & MambaMorph \citep{wang2025mamba} & 0.286$\pm$0.282 & 1.93$\pm$0.34 & 0.701$\pm$0.053 & 0.715$\pm$0.060 & 0.727$\pm$0.068 & 0.752$\pm$0.045 & 72.51$\pm$5.30 \\
    & TransMorph-rigid  & 0.557$\pm$0.361 & \underline{1.52$\pm$0.31} & \underline{0.764$\pm$0.047} & \underline{0.796$\pm$0.050} & \underline{0.788$\pm$0.053} & \underline{0.800$\pm$0.054} & \underline{78.71$\pm$4.49} \\
    & MambaMorph-rigid  & 0.889$\pm$0.636 & 1.65$\pm$0.28 & 0.728$\pm$0.054 & 0.759$\pm$0.049 & 0.765$\pm$0.035 & 0.782$\pm$0.031 & 75.84$\pm$3.42 \\
    & MSR(Ours) & 0.151$\pm$0.143 & \textbf{1.43$\pm$0.28} & \textbf{0.796$\pm$0.041} & \textbf{0.808$\pm$0.054} & \textbf{0.796$\pm$0.041} & \textbf{0.817$\pm$0.036} & \textbf{79.84$\pm$3.59} \\
    \hline
    \multirow{8}{*}{HN} & Affine & -- & 2.53$\pm$1.31 & 0.656$\pm$0.194 & 0.682$\pm$0.189 & 0.617$\pm$0.241 & 0.592$\pm$0.255 & 63.68$\pm$21.64 \\
    & SyN \citep{avants2008symmetric} & 0.290$\pm$0.412 & 2.36$\pm$0.86 & 0.725$\pm$0.111 & 0.770$\pm$0.092 & 0.695$\pm$0.093 & 0.680$\pm$0.108 & 71.73$\pm$9.60 \\
    & VoxelMorph \citep{balakrishnan2019voxelmorph} & \textbf{<0.0001} & 0.83$\pm$0.47 & 0.977$\pm$0.014 & 0.943$\pm$0.030 & 0.894$\pm$0.062 & 0.894$\pm$0.064 & 92.76$\pm$4.14 \\
    & TransMorph \citep{chen2022transmorph} & 0.034$\pm$0.067 & 0.88$\pm$0.16 & 0.981$\pm$0.007 & 0.959$\pm$0.019 & 0.924$\pm$0.043 & 0.911$\pm$0.041 & 94.41$\pm$2.63 \\
    & MambaMorph \citep{wang2025mamba} & \underline{0.027$\pm$0.043} & 0.96$\pm$0.14 & 0.973$\pm$0.006 & 0.960$\pm$0.013 & 0.929$\pm$0.036 & 0.911$\pm$0.040 & 94.35$\pm$2.32 \\
    & TransMorph-rigid  & 0.042$\pm$0.069 & \textbf{0.72$\pm$0.15} & \textbf{0.983$\pm$0.003} & \underline{0.975$\pm$0.008} & \underline{0.950$\pm$0.017} & \underline{0.940$\pm$0.019} & \underline{96.23$\pm$1.11} \\
    & MambaMorph-rigid  & \textbf{<0.0001} & 0.84$\pm$0.12 & \underline{0.981$\pm$0.005} & 0.967$\pm$0.013 & 0.948$\pm$0.029 & 0.937$\pm$0.026 & 95.85$\pm$1.69 \\
    & MSR(Ours) & 0.036$\pm$0.057 & \underline{0.81$\pm$0.17} & \underline{0.981$\pm$0.009} & \textbf{0.976$\pm$0.008} & \textbf{0.952$\pm$0.018} & \textbf{0.946$\pm$0.017} & \textbf{96.43$\pm$1.13} \\
    \hline
  \end{tabular*}}
  \setlength{\extrarowheight}{0pt}
  \renewcommand{\arraystretch}{1}
  }
\end{table*}

\begin{figure*}[t]
  \centering
  \includegraphics[width=0.72\textwidth]{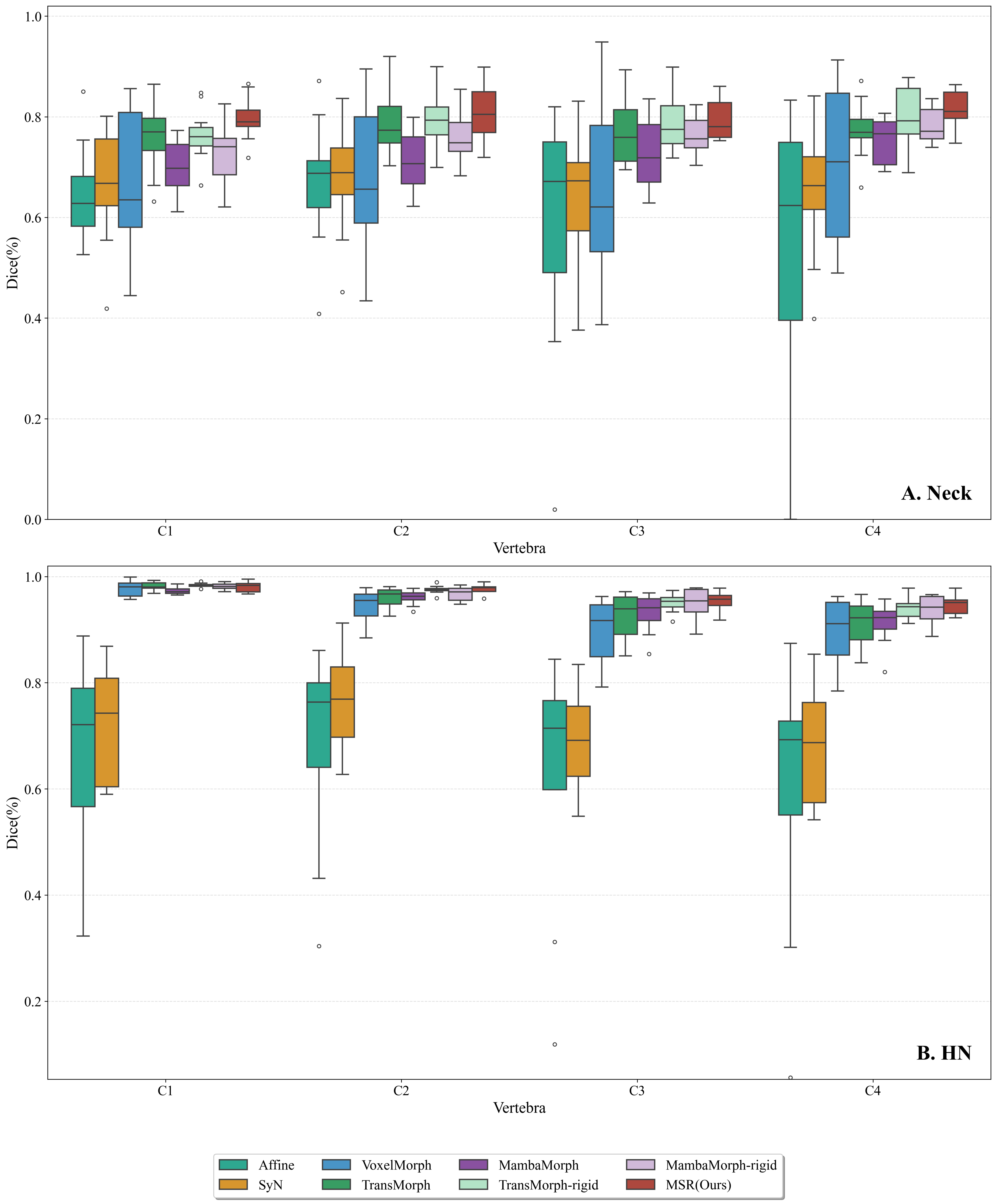}
  \caption{Boxplots of Dice scores for vertebra structures on Neck and HN. The central horizontal line in each box represents the median, while the bottom and top edges of the box indicate the 25th (Q1) and 75th (Q3) percentiles, respectively. The whiskers extend to the maximum and minimum values within the interquartile range (IQR) from the box edges. Data points plotted individually as dots beyond the whiskers represent outliers, indicating cases with atypical registration performance.}
  \label{fig:dice-boxplots-neck-hn}
\end{figure*}

As shown in \tabref{tab:neck-hn-comparison} and \figref{fig:dice-boxplots-neck-hn}, the proposed MSR achieves the best or near-best performance on both datasets and demonstrates consistent advantages across multiple evaluation metrics. On the Neck dataset, MSR attains the best HD95 of $1.43 \pm 0.28$ among all methods. In terms of Dice, MSR achieves $0.796 \pm 0.041$, $0.808 \pm 0.054$, $0.796 \pm 0.041$, and $0.817 \pm 0.036$ on C1--C4, respectively, all of which are the best results, yielding an overall average Dice of $79.84 \pm 3.59\%$.

On the HN dataset, owing to the relatively good initial alignment, the performance differences are less pronounced; nevertheless, MSR still achieves the highest average Dice of $96.43 \pm 1.13\%$. Specifically, in the C2--C4 regions, MSR obtains $0.976 \pm 0.008$, $0.952 \pm 0.018$, and $0.946 \pm 0.017$, respectively, outperforming the other methods. Overall, MSR achieves superior performance across Dice, HD95, and Jacobian metrics, demonstrating more stable and reliable registration results. This can be attributed to the integration of the rigid module into the deformable framework, which effectively satisfies the requirements of rigid--deformable registration, as well as the MSL module in the deformable stage, enabling MSR to jointly capture global and local structural information.

The rigid registration module in MSR improves registration performance while preserving the rigidity of bony structures. As shown in \tabref{tab:neck-hn-comparison}, comparisons between TransMorph and TransMorph-rigid, as well as MambaMorph and MambaMorph-rigid, indicate that incorporating the rigid module consistently improves Dice scores on both datasets. On the Neck dataset, the average Dice of TransMorph-rigid increases from $77.53\%$ to $78.71\%$, and that of MambaMorph-rigid improves from $72.51\%$ to $75.84\%$. On the HN dataset, TransMorph-rigid improves from $94.41\%$ to $96.23\%$, while MambaMorph-rigid increases from $94.35\%$ to $95.85\%$.

\begin{figure*}[t]
  \centering
  \includegraphics[width=0.96\textwidth]{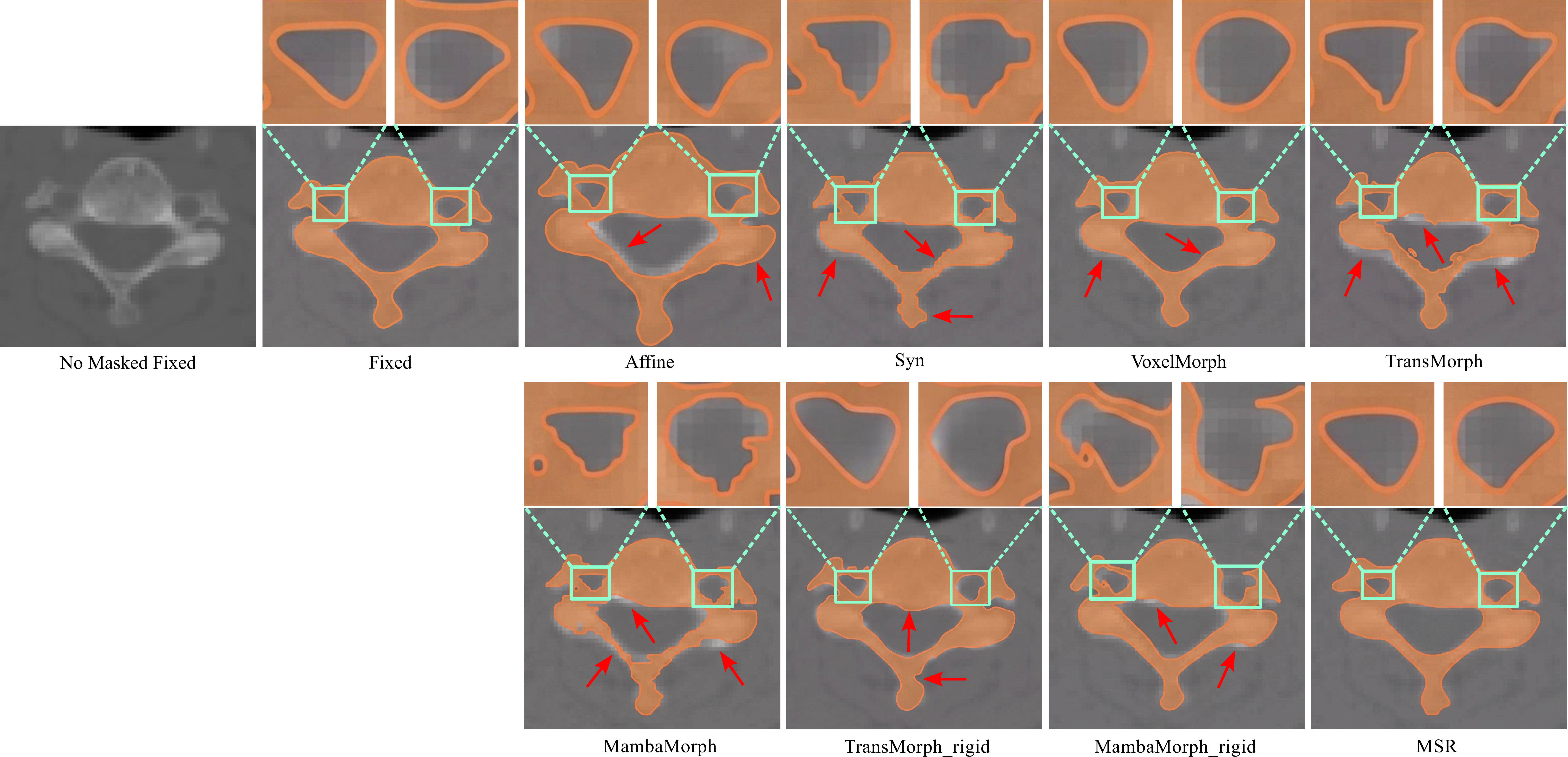}
  \caption{Qualitative comparison of different methods on the Neck dataset. For each method, the overlaid segmentation masks of the fixed and registered moving images are shown on the axial slice of the C4 vertebra. The first two rows present the deformed masks overlaid on CT, while the last two rows show the deformed masks overlaid on the registered MRI. Red arrows indicate unrealistic distortions of the masks, and red boxes highlight the transverse foramen region through which the vertebral artery passes. The numbers in the images denote the corresponding Dice scores.}
  \label{fig:neck-qualitative}
\end{figure*}

Furthermore, as illustrated in \figref{fig:neck-qualitative}, qualitative comparisons on the C4 vertebra in the Neck dataset show that conventional methods (e.g., Affine and SyN \citep{avants2008symmetric}) and purely deformable models (e.g., VoxelMorph \citep{balakrishnan2019voxelmorph}, TransMorph \citep{chen2022transmorph}, and MambaMorph \citep{wang2025mamba}) still exhibit noticeable structural misalignment and unrealistic deformations after registration, as indicated by the red arrows. After incorporating the rigid registration module into TransMorph and MambaMorph, these structural inconsistencies are effectively alleviated.

Furthermore, building upon the rigid registration module, the introduction of the MSL module leads to additional performance improvements. As shown in \tabref{tab:neck-hn-comparison}, MSR consistently outperforms both TransMorph-rigid and MambaMorph-rigid across the two datasets. On the Neck dataset, MSR achieves a 1.13\% improvement in average Dice compared to the second-best method, TransMorph-rigid, with the most notable gain observed in the anatomically complex C1 region. In addition, regarding the ``\% of $|J| < 0$'' metric, MSR ($0.151 \pm 0.143$) is significantly lower than MambaMorph-rigid, indicating improved deformation regularity. On the HN dataset, MSR attains an average Dice of 96.43\%, surpassing TransMorph-rigid (96.23\%) and MambaMorph-rigid (95.85\%). These results demonstrate that the MSL module effectively models both global contextual information and local fine-grained details in a multi-scale manner, thereby enhancing the representation of complex anatomical structures.

\begin{figure*}[t]
  \centering
  \includegraphics[width=0.96\textwidth]{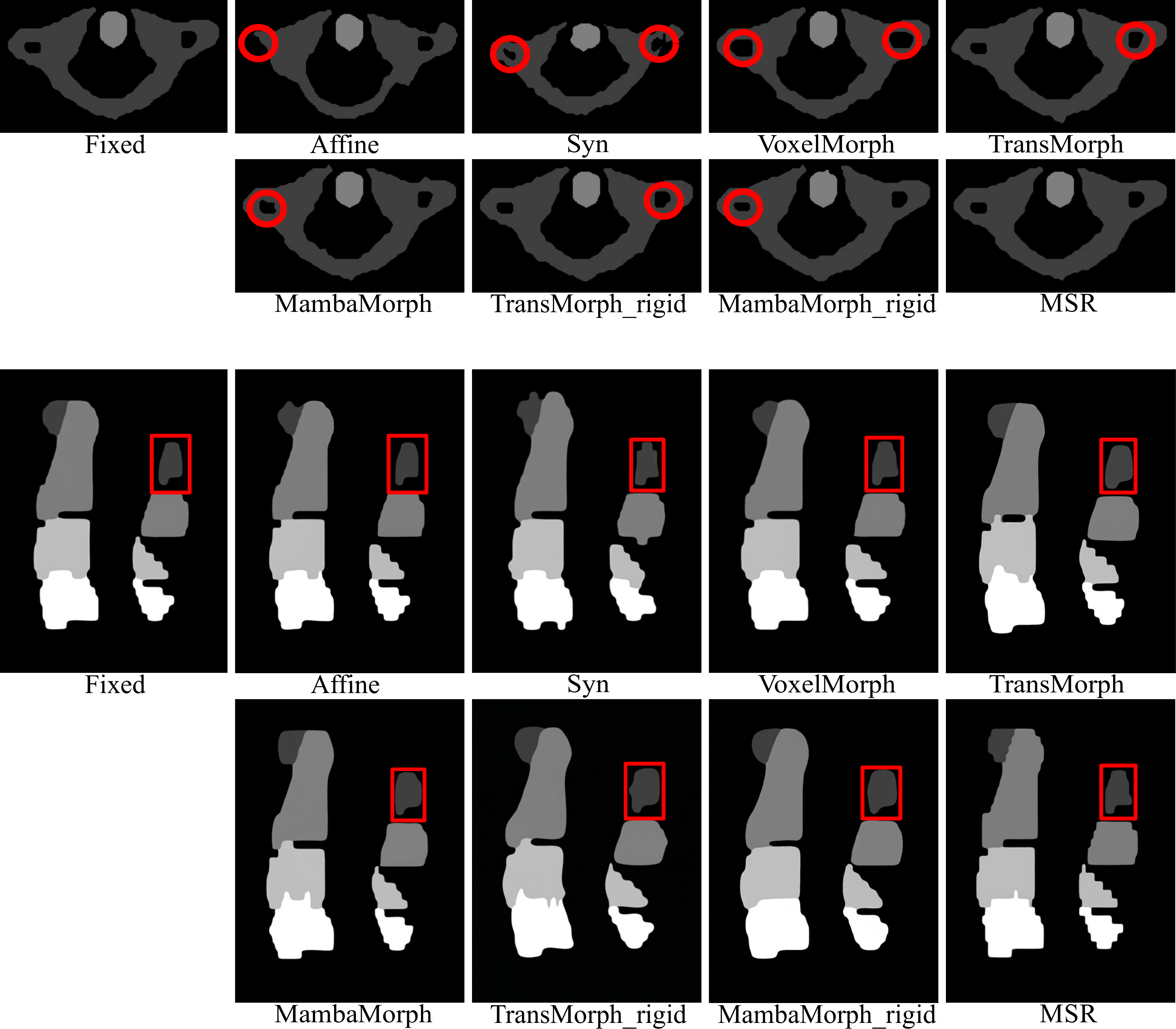}
  \caption{Qualitative comparison of different methods on the HN dataset. For each method, the mask deformation of the C1 structure is shown before and after registration. The first two rows present the mask deformation results on axial slices, while the last two rows show the corresponding results on sagittal slices. Red circles indicate local abnormal deformations in the transverse foramen region through which the vertebral artery passes, and red boxes highlight structural shape changes of the mask on the sagittal view of C1.}
  \label{fig:hn-qualitative}
\end{figure*}

Moreover, as illustrated in \figref{fig:neck-qualitative}, although the rigid module in TransMorph-rigid and MambaMorph-rigid alleviates global structural misalignment, boundary blurring and local deformation artifacts still persist in critical fine structures, such as the transverse foramen (highlighted by red boxes). In contrast, after incorporating the MSL module, MSR preserves more complete structural morphology and sharper boundaries in these regions, resulting in more stable registration outcomes. As shown in \figref{fig:hn-qualitative}, although TransMorph-rigid achieves slightly higher performance in the C1 region, it still exhibits local unrealistic deformations around the transverse foramen---where the vertebral artery passes---marked by red circles, indicating insufficient physical plausibility in local structures. Further observations from the sagittal view reveal that MSR, with the MSL module, maintains anatomically consistent and physically plausible deformation patterns in the C1 region.

\subsubsection{Comparative study on TH}

\begin{table*}[t]
  \makeatletter
  \long\def\@makecaption#1#2{%
    \centering
    \vskip 6pt
    {\parbox{\textwidth}{\rightskip=0pt\fontfamily{ptm}\selectfont\fontsize{10pt}{12pt}\selectfont\textbf{#1}\par#2\par\vskip4pt}}%
  }
  \makeatother
  \caption{Quantitative comparison on the TH test dataset. Results are reported as mean $\pm$ standard deviation (Mean $\pm$ Std). The best performance is highlighted in \textbf{bold}, and the second-best is \underline{underlined}. An upward arrow ($\uparrow$) indicates that higher values are better, while a downward arrow ($\downarrow$) indicates that lower values are better.}
  \label{tab:th-comparison}
  \centering
  {\fontfamily{ptm}\selectfont
  \fontsize{8pt}{9.8pt}\selectfont
  \renewcommand{\arraystretch}{1.18}
  \setlength{\tabcolsep}{3pt}
  \setlength{\extrarowheight}{1pt}
  \resizebox{\textwidth}{!}{%
  \begin{tabular*}{1.03\textwidth}{@{\hspace{4pt}\extracolsep{\fill}}lccccccc@{\hspace{4pt}}}
    \hline
    Methods & \% of $|J| < 0$ $\downarrow$ & HD95 $\downarrow$ & T6 $\uparrow$ & T7 $\uparrow$ & T8 $\uparrow$ & T9 $\uparrow$ & Average Dice $\uparrow$ \\
    \hline
    Affine & -- & 2.14$\pm$0.97 & 0.739$\pm$0.166 & 0.742$\pm$0.152 & 0.754$\pm$0.133 & 0.748$\pm$0.124 & 74.61 $\pm$ 13.97 \\
    SyN \citep{avants2008symmetric} & 0.027$\pm$0.067 & 2.14$\pm$1.01 & 0.763$\pm$0.168 & 0.779$\pm$0.143 & 0.797$\pm$0.114 & 0.795$\pm$0.101 & 78.39 $\pm$ 12.72 \\
    VoxelMorph \citep{balakrishnan2019voxelmorph} & \textbf{<0.0001} & 1.59$\pm$0.85 & 0.862$\pm$0.080 & 0.861$\pm$0.077 & 0.861$\pm$0.078 & 0.870$\pm$0.068 & 86.38$\pm$7.24 \\
    TransMorph \citep{chen2022transmorph} & 0.122$\pm$0.257 & 1.48$\pm$0.53 & 0.862$\pm$0.064 & 0.864$\pm$0.062 & 0.864$\pm$0.061 & 0.874$\pm$0.053 & 86.65 $\pm$ 5.51 \\
    MambaMorph \citep{wang2025mamba} & \underline{0.066$\pm$0.128} & 1.43$\pm$0.47 & 0.854$\pm$0.054 & 0.856$\pm$0.055 & 0.865$\pm$0.058 & 0.880$\pm$0.045 & 86.43 $\pm$ 4.94 \\
    TransMorph-rigid  & 1.669$\pm$1.345 & \textbf{1.31$\pm$0.37} & \underline{0.867$\pm$0.044} & \underline{0.871$\pm$0.041} & 0.874$\pm$0.037 & \textbf{0.887$\pm$0.027} & \underline{87.54 $\pm$ 3.54} \\
    MambaMorph-rigid  & 0.953$\pm$3.711 & 1.40$\pm$0.40 & 0.862$\pm$0.047 & 0.870$\pm$0.040 & \underline{0.880$\pm$0.043} & 0.875$\pm$0.038 & 87.17 $\pm$ 3.53 \\
    MSR(Ours) & 0.221$\pm$0.326 & \underline{1.33$\pm$0.45} & \textbf{0.875$\pm$0.067} & \textbf{0.884$\pm$0.048} & \textbf{0.889$\pm$0.042} & \underline{0.883$\pm$0.040} & \textbf{88.30$\pm$4.62} \\
    \hline
  \end{tabular*}}
  \setlength{\extrarowheight}{0pt}
  \renewcommand{\arraystretch}{1}
  }
\end{table*}

Unlike the cervical spine regions in the Neck and HN datasets, the TH dataset focuses on thoracic vertebrae, which exhibit notable differences in anatomical morphology and spatial distribution. Under this cross-anatomical setting, the experimental results provide an effective evaluation of model generalization. As shown in \tabref{tab:th-comparison}, MSR still achieves the best overall performance on the TH dataset. Specifically, MSR attains an average Dice of $88.30 \pm 4.62\%$, outperforming the second-best methods, TransMorph-rigid ($87.54 \pm 3.54\%$) and MambaMorph-rigid ($87.17 \pm 3.53\%$). For individual vertebrae T6--T9, MSR achieves Dice scores of $0.875 \pm 0.067$, $0.884 \pm 0.048$, $0.889 \pm 0.042$, and $0.883 \pm 0.040$, respectively, all of which are the best or near-best results, demonstrating stable performance across different vertebral levels. In terms of HD95, MSR achieves $1.33 \pm 0.45$, outperforming most competing methods. Overall, MSR maintains superior performance on both Dice and HD95, indicating strong generalization capability.

\begin{figure*}[t]
  \centering
  \includegraphics[width=0.96\textwidth]{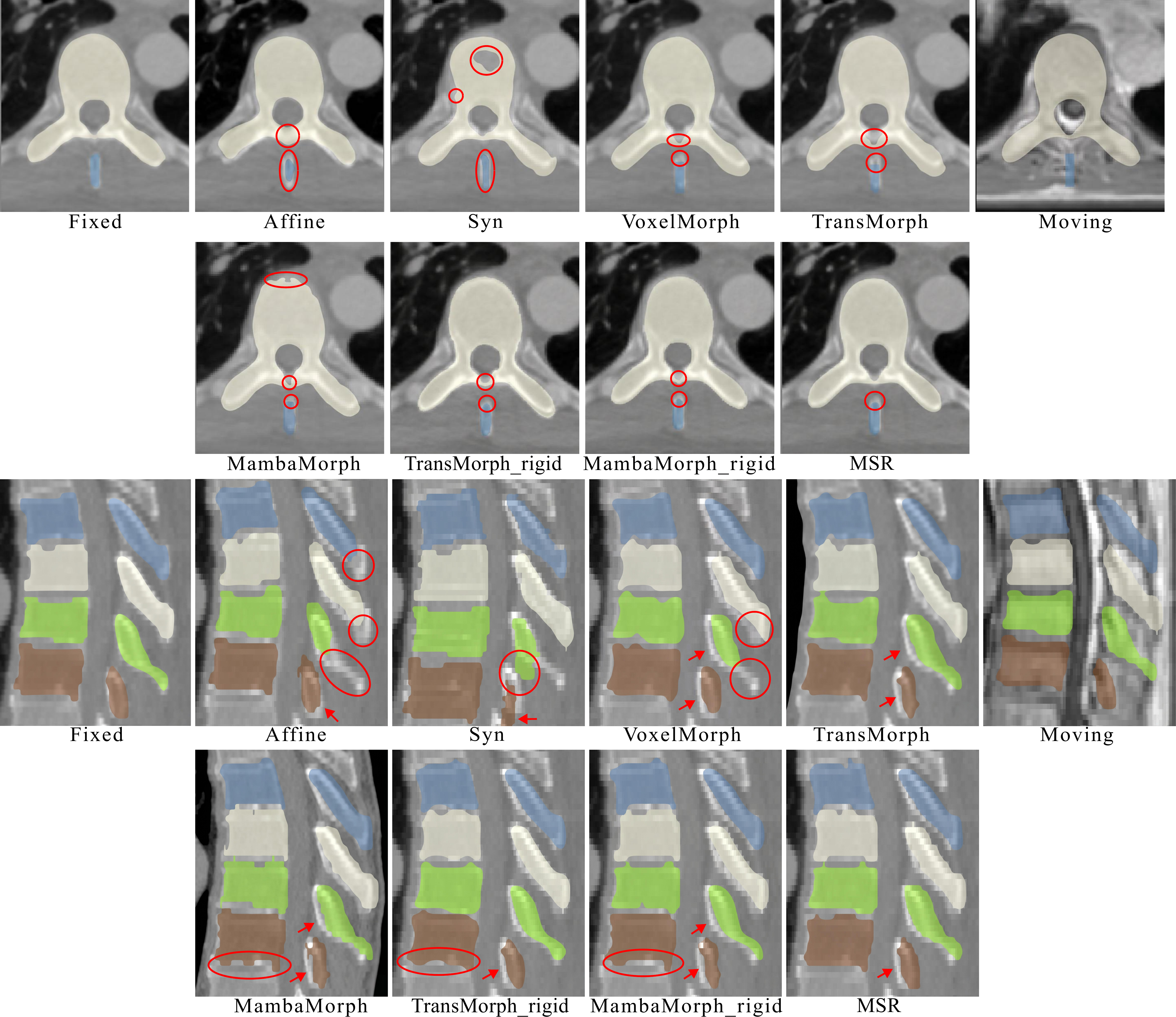}
  \caption{Qualitative comparison of different methods on the TH dataset. For each method, the mask deformation of the T7 structure is shown before and after registration. The first two rows present the mask deformation results on axial slices, while the last two rows show the corresponding results on sagittal slices. Red circles indicate local abnormal deformations, and red arrows highlight structural displacements of the mask on the sagittal view.}
  \label{fig:th-qualitative}
\end{figure*}

As observed in \tabref{tab:th-comparison}, after incorporating the rigid registration module, the average Dice of TransMorph and MambaMorph improves from 86.65\% and 86.43\% to 87.54\% and 87.17\%, respectively. This indicates that the rigid module provides effective rigidity constraints, thereby reducing unrealistic deformations introduced by purely deformable modeling. Consistent with this observation, the sagittal views in \figref{fig:th-qualitative} show that the incorporation of the rigid module significantly alleviates structural distortions of the vertebrae. This further demonstrates that the rigid registration module effectively enforces anatomical rigidity and improves the overall plausibility of the deformation.

\begin{figure*}[t]
  \centering
  \includegraphics[width=0.96\textwidth]{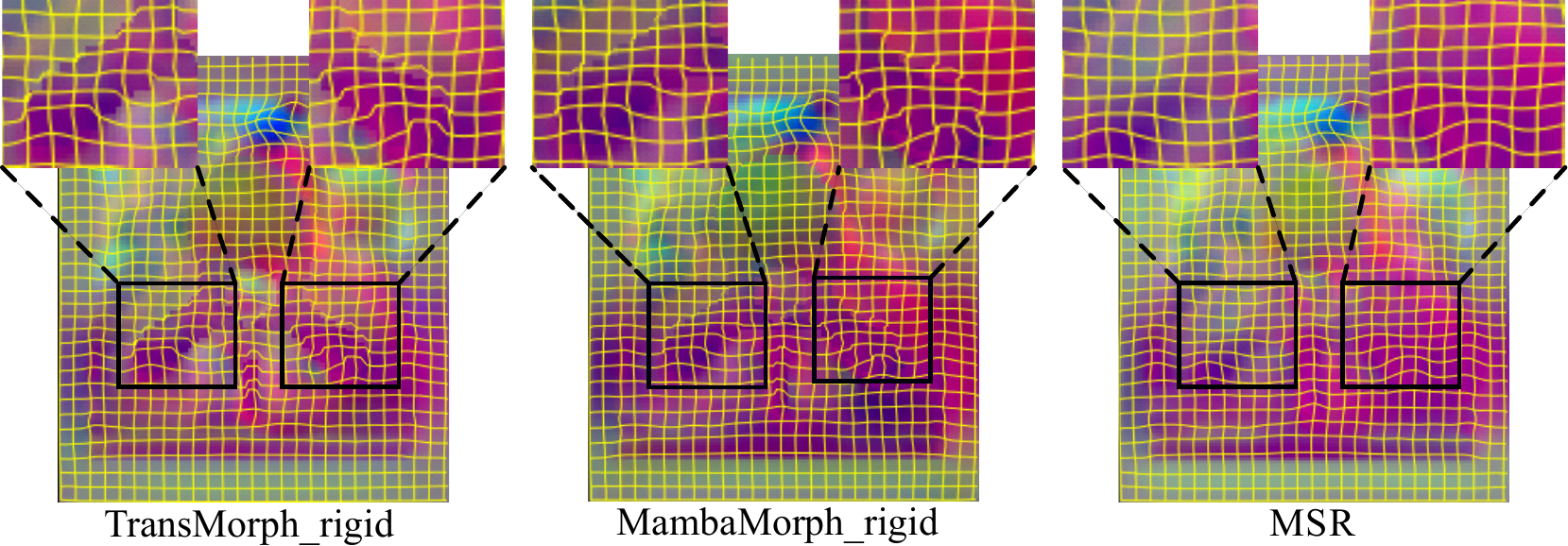}
  \caption{Zoomed-in visualization of deformation fields in selected boundary regions, used to compare the deformation behavior of different methods near structural boundaries.}
  \label{fig:flow-boundary}
\end{figure*}

As shown in \tabref{tab:th-comparison}, on top of the rigid registration module, the introduction of the MSL module further improves the average Dice of MSR over TransMorph-rigid and MambaMorph-rigid by 0.76\% and 1.13\%, respectively. Meanwhile, MSR achieves lower values in the Jacobian determinant ($J_{\phi}$) compared to both methods. This can be attributed to the ability of MSL to jointly capture global contextual information and local fine-grained features across different spatial scales, thereby enhancing the representation of complex structures while preserving topological plausibility. As illustrated in \figref{fig:th-qualitative}, the axial views show that in regions with fine anatomical details, such as the laminar junction, TransMorph-rigid and MambaMorph-rigid still suffer from blurred details or structural degradation, whereas MSR preserves clearer structural boundaries and more complete anatomical morphology. Furthermore, as observed in \figref{fig:flow-boundary}, the deformation fields generated by MSR are smoother and more continuous in boundary regions, without evident local distortions or irregular warping.

\subsubsection{Results on external test dataset}

To further evaluate the generalization capability of the proposed method, cross-dataset experiments are conducted as shown in \tabreftwo{tab:cross-neck-to-hn}{tab:cross-hn-to-neck}. Two settings are considered: (1) training on the Neck dataset and testing on the HN dataset, and (2) training on the HN dataset and testing on the Neck dataset. Owing to differences in imaging devices, acquisition parameters, and anatomical distributions between the training and testing data, this experimental setup provides an effective assessment of model robustness under distribution shifts.

\begin{table*}[t]
  \makeatletter
  \long\def\@makecaption#1#2{%
    \centering
    \vskip 6pt
    {\parbox{\textwidth}{\rightskip=0pt\fontfamily{ptm}\selectfont\fontsize{10pt}{12pt}\selectfont\textbf{#1}\par#2\par\vskip4pt}}%
  }
  \makeatother
  \caption{Dice scores of different registration methods for vertebrae on the HN dataset, using models trained on the Neck dataset. Results are reported as mean $\pm$ standard deviation (Mean $\pm$ Std). The best performance is highlighted in \textbf{bold}, and the second-best is \underline{underlined}. An upward arrow ($\uparrow$) indicates that higher values are better.}
  \label{tab:cross-neck-to-hn}
  \centering
  {\fontfamily{ptm}\selectfont
  \fontsize{8pt}{9.8pt}\selectfont
  \renewcommand{\arraystretch}{1.18}
  \setlength{\tabcolsep}{3pt}
  \setlength{\extrarowheight}{1pt}
  \resizebox{0.82\textwidth}{!}{%
  \begin{tabular*}{0.90\textwidth}{@{\hspace{4pt}\extracolsep{\fill}}lccccc@{\hspace{4pt}}}
    \hline
    Methods & C1 $\uparrow$ & C2 $\uparrow$ & C3 $\uparrow$ & C4 $\uparrow$ & Average Dice $\uparrow$ \\
    \hline
    TransMorph \citep{chen2022transmorph} & 0.819$\pm$0.039 & 0.841$\pm$0.031 & 0.819$\pm$0.025 & 0.817$\pm$0.031 & 82.42$\pm$2.74 \\
    MambaMorph \citep{wang2025mamba} & 0.831$\pm$0.041 & 0.837$\pm$0.034 & 0.811$\pm$0.027 & 0.800$\pm$0.051 & 82.01$\pm$3.58 \\
    TransMorph-rigid  & \textbf{0.869$\pm$0.033} & \textbf{0.896$\pm$0.026} & \textbf{0.907$\pm$0.032} & \textbf{0.887$\pm$0.041} & \textbf{88.98$\pm$2.73} \\
    MambaMorph-rigid  & 0.813$\pm$0.024 & 0.853$\pm$0.023 & \underline{0.843$\pm$0.035} & \underline{0.851$\pm$0.042} & \underline{84.03$\pm$2.72} \\
    MSR(Ours) & 0.811$\pm$0.015 & \underline{0.867$\pm$0.016} & 0.833$\pm$0.019 & 0.838$\pm$0.022 & 83.71$\pm$1.56 \\
    \hline
  \end{tabular*}}
  \setlength{\extrarowheight}{0pt}
  \renewcommand{\arraystretch}{1}
  }
\end{table*}

When MSR is trained on the Neck dataset and directly applied to the HN dataset, the performance of all methods degrades, indicating a clear cross-modality and cross-region distribution gap. MSR achieves $83.71 \pm 1.56\%$, outperforming TransMorph ($82.42 \pm 2.74\%$) and MambaMorph ($82.01 \pm 3.58\%$). Across individual structures (C1--C4), MSR exhibits more balanced performance and a smaller standard deviation ($\pm 1.56$), demonstrating better stability and robustness under cross-dataset settings.

\begin{table*}[t]
  \makeatletter
  \long\def\@makecaption#1#2{%
    \centering
    \vskip 6pt
    {\parbox{\textwidth}{\rightskip=0pt\fontfamily{ptm}\selectfont\fontsize{10pt}{12pt}\selectfont\textbf{#1}\par#2\par\vskip4pt}}%
  }
  \makeatother
  \caption{Dice scores of different registration methods for vertebrae on the Neck dataset, using models trained on the HN dataset. Results are reported as mean $\pm$ standard deviation (Mean $\pm$ Std). The best performance is highlighted in \textbf{bold}, and the second-best is \underline{underlined}. An upward arrow ($\uparrow$) indicates that higher values are better.}
  \label{tab:cross-hn-to-neck}
  \centering
  {\fontfamily{ptm}\selectfont
  \fontsize{8pt}{9.8pt}\selectfont
  \renewcommand{\arraystretch}{1.18}
  \setlength{\tabcolsep}{3pt}
  \setlength{\extrarowheight}{1pt}
  \resizebox{0.82\textwidth}{!}{%
  \begin{tabular*}{0.90\textwidth}{@{\hspace{4pt}\extracolsep{\fill}}lccccc@{\hspace{4pt}}}
    \hline
    Methods & C1 $\uparrow$ & C2 $\uparrow$ & C3 $\uparrow$ & C4 $\uparrow$ & Average Dice $\uparrow$ \\
    \hline
    TransMorph \citep{chen2022transmorph} & 0.649$\pm$0.013 & 0.644$\pm$0.014 & 0.622$\pm$0.020 & 0.630$\pm$0.020 & 63.63$\pm$1.50 \\
    MambaMorph \citep{wang2025mamba} & 0.671$\pm$0.013 & 0.674$\pm$0.016 & 0.669$\pm$0.023 & 0.673$\pm$0.019 & 67.18$\pm$1.31 \\
    TransMorph-rigid  & 0.676$\pm$0.011 & 0.680$\pm$0.012 & 0.678$\pm$0.011 & 0.681$\pm$0.008 & 67.90$\pm$0.91 \\
    MambaMorph-rigid  & \textbf{0.692$\pm$0.005} & \underline{0.687$\pm$0.008} & \underline{0.684$\pm$0.013} & \underline{0.683$\pm$0.014} & \underline{68.66$\pm$0.82} \\
    MSR(Ours) & \underline{0.691$\pm$0.004} & \textbf{0.694$\pm$0.003} & \textbf{0.693$\pm$0.005} & \textbf{0.690$\pm$0.011} & \textbf{69.24$\pm$0.45} \\
    \hline
  \end{tabular*}}
  \setlength{\extrarowheight}{0pt}
  \renewcommand{\arraystretch}{1}
  }
\end{table*}

When trained on the HN dataset and evaluated on the Neck dataset, the overall performance further decreases because of the more complex joint structures and stronger non-rigid deformations in the Neck data. Under this more challenging setting, MSR achieves the highest average Dice of $69.24 \pm 0.45\%$, outperforming all competing methods, including TransMorph-rigid ($67.90 \pm 0.91\%$) and MambaMorph-rigid ($68.66 \pm 0.82\%$). Moreover, MSR maintains the best or near-best performance across all C1--C4 structures, with the lowest variance ($\pm 0.45$), indicating strong generalization capability under complex deformation scenarios and significant domain shifts.

\subsubsection{Ablation Study}

\begin{table*}[t]
  \makeatletter
  \long\def\@makecaption#1#2{%
    \centering
    \vskip 6pt
    {\parbox{\textwidth}{\rightskip=0pt\fontfamily{ptm}\selectfont\fontsize{10pt}{12pt}\selectfont\textbf{#1}\par#2\par\vskip4pt}}%
  }
  \makeatother
  \caption{Ablation study on the rigid registration module. Results are reported as mean $\pm$ standard deviation (Mean $\pm$ Std). The best performance is highlighted in \textbf{bold}. An upward arrow ($\uparrow$) indicates that higher values are better.}
  \label{tab:ablation-rigid}
  \centering
  {\fontfamily{ptm}\selectfont
  \fontsize{8pt}{9.8pt}\selectfont
  \renewcommand{\arraystretch}{1.18}
  \setlength{\tabcolsep}{3pt}
  \setlength{\extrarowheight}{1pt}
  \resizebox{0.82\textwidth}{!}{%
  \begin{tabular*}{0.90\textwidth}{@{\hspace{4pt}\extracolsep{\fill}}llccccc@{\hspace{4pt}}}
    \hline
    Datasets & Methods & C1/T6 $\uparrow$ & C2/T7 $\uparrow$ & C3/T8 $\uparrow$ & C4/T9 $\uparrow$ & Average Dice $\uparrow$ \\
    \hline
    \multirow{2}{*}{Neck} & MS & 0.766$\pm$0.052 & 0.779$\pm$0.054 & 0.766$\pm$0.051 & 0.777$\pm$0.045 & 77.20$\pm$4.57 \\
    & MSR & \textbf{0.796$\pm$0.041} & \textbf{0.808$\pm$0.054} & \textbf{0.796$\pm$0.041} & \textbf{0.817$\pm$0.036} & \textbf{79.84$\pm$3.59} \\
    \hline
    \multirow{2}{*}{HN} & MS & 0.976$\pm$0.006 & 0.967$\pm$0.012 & 0.939$\pm$0.029 & 0.919$\pm$0.038 & 95.07$\pm$2.01 \\
    & MSR & \textbf{0.981$\pm$0.009} & \textbf{0.976$\pm$0.008} & \textbf{0.952$\pm$0.018} & \textbf{0.946$\pm$0.017} & \textbf{96.43$\pm$1.13} \\
    \hline
    \multirow{2}{*}{TH} & MS & 0.867$\pm$0.051 & 0.870$\pm$0.051 & 0.870$\pm$0.048 & 0.875$\pm$0.042 & 87.06$\pm$4.44 \\
    & MSR & \textbf{0.875$\pm$0.067} & \textbf{0.884$\pm$0.048} & \textbf{0.889$\pm$0.042} & \textbf{0.883$\pm$0.040} & \textbf{88.30$\pm$4.62} \\
    \hline
  \end{tabular*}}
  \setlength{\extrarowheight}{0pt}
  \renewcommand{\arraystretch}{1}
  }
\end{table*}

To validate the effectiveness of the rigid registration module, we remove the rigid branch from the full model (MSR), retaining only the deformable registration component, denoted as MS, and conduct comparative experiments on the Neck, HN, and TH datasets. As shown in \tabref{tab:ablation-rigid}, incorporating the rigid module consistently improves performance across all three datasets. On the Neck dataset, the average Dice increases significantly from $77.20 \pm 4.57\%$ to $79.84 \pm 3.59\%$, corresponding to an improvement of approximately 2.6 percentage points, with consistent gains across all C1--C4 structures. On the HN dataset, where performance is already close to saturation, MSR still improves the average Dice from $95.07 \pm 2.01\%$ to $96.43 \pm 1.13\%$. Notably, the standard deviation is substantially reduced, indicating that the rigid module primarily enhances stability and consistency in this scenario. On the TH dataset, the average Dice improves from $87.06 \pm 4.44\%$ to $88.30 \pm 4.62\%$. These results demonstrate that the rigid registration module is a key component for improving both performance and robustness of the model.

\section{Discussions and conclusions}

This study addresses the challenge of jointly modeling rigid structures and deformable motion in cervical CT--MRI registration by proposing a structure-aware rigid--deformable hybrid framework, termed MSR. Conventional rigid or piecewise methods struggle to capture complex local deformations, whereas existing deep learning approaches predominantly rely on purely deformable modeling, often neglecting the rigidity constraints of bony structures and thus compromising structural consistency. To overcome this limitation, we explicitly model rigid deformation and integrate it with deformable fields within a unified framework. In addition, we introduce the MSL module to enhance both global and local feature representations. Furthermore, we construct and release the R-D-Reg cervical CT--MRI registration dataset, providing a standardized benchmark for this task.

Traditional spine registration methods typically rely on vertebral masks to constrain different anatomical structures, enabling separate modeling of multiple vertebrae and surrounding soft-tissue deformation fields. However, as shown by the results of Affine and SyN \citep{avants2008symmetric} in \tabreftwo{tab:neck-hn-comparison}{tab:th-comparison}, their performance degrades significantly without mask constraints. Qualitative results in \figreftwo{fig:neck-qualitative}{fig:hn-qualitative} further reveal that Affine mainly produces large global misalignments, whereas SyN \citep{avants2008symmetric} introduces unrealistic distortions within bony structures. Such artifacts compromise anatomical consistency and reduce the overall registration quality.

Most current deep learning-based registration methods focus on global deformable modeling. Although they often achieve comparable or even superior accuracy compared with traditional approaches, as shown in \tabreftwo{tab:neck-hn-comparison}{tab:th-comparison}, VoxelMorph \citep{balakrishnan2019voxelmorph}, TransMorph \citep{chen2022transmorph}, and MambaMorph \citep{wang2025mamba} yield relatively low Dice scores in cervical CT--MRI registration. Notably, while VoxelMorph achieves the lowest values in the ``\% of $|J| < 0$'' metric, its registration accuracy remains inferior to that of the proposed method. This suggests that purely deformable approaches are insufficient for this task and highlights the need for more effective modeling strategies.

Based on these observations, we first introduce a rigid registration module and extend two strong deformable models, TransMorph \citep{chen2022transmorph} and MambaMorph \citep{wang2025mamba}, into TransMorph-rigid and MambaMorph-rigid. These extended models explicitly generate rigid deformation fields and fuse them with deformable fields. As shown in \tabreftwo{tab:neck-hn-comparison}{tab:th-comparison}, both variants achieve significant improvements in Dice compared with their original counterparts. This demonstrates that the proposed rigid module effectively enhances the alignment of rigid structures without compromising deformable modeling capability. From a physical perspective, vertebrae behave as rigid bodies whose motion is constrained to rotation and translation. In contrast, purely deformable methods, lacking such priors, may approximate rigid regions using high degrees of freedom, leading to non-physical distortions. The proposed rigid module imposes low-degree-of-freedom rigid transformations for each vertebra in the feature space, thereby reducing the complexity of the deformation space. Moreover, a mask-guided local modeling strategy is adopted, allowing each vertebra to independently estimate its rigid motion. This design is consistent with the biomechanical characteristics of the cervical spine, where different vertebrae exhibit relatively independent motion patterns. Compared with a single global rigid transformation, this decomposition enables more accurate modeling of complex local pose variations and avoids incorrectly compressing multiple relative motions into a single transformation.

Furthermore, we observe that directly incorporating a rigid module into models such as TransMorph \citep{chen2022transmorph} and MambaMorph \citep{wang2025mamba} may still introduce unrealistic deformations in regions with fine anatomical structures, such as the transverse foramen along the vertebral artery pathway, as shown in \figref{fig:neck-qualitative}. This issue arises because existing methods often rely on a single modeling mechanism to simultaneously capture large-scale structural consistency and fine-grained local alignment, leading to a trade-off between the two. To address this, we introduce the MSL module in the deformable registration stage. The Mamba-based global modeling branch captures long-range dependencies and enforces structural consistency across regions, such as spinal alignment and inter-vertebral spatial relationships. In contrast, the Swin Transformer branch focuses on fine-grained local alignment, particularly for small anatomical details such as the transverse foramen. The MSL module further employs a gating mechanism to adaptively balance global and local information. As shown in \figreftwo{fig:neck-qualitative}{fig:hn-qualitative}, this design effectively preserves the structure of the transverse foramen; additionally, as illustrated in \figref{fig:th-qualitative}, it achieves more anatomically consistent modeling in regions such as the laminar junction in the thoracic spine. These results collectively validate the effectiveness of the proposed design. Overall, MSR achieves accurate and robust CT--MRI registration in the cervical spine region.

Despite these promising results, several limitations remain. First, although current deformable registration models perform well in soft-tissue regions, the lack of precise soft-tissue annotations prevents a comprehensive evaluation of soft-tissue alignment using Dice. Second, traditional rigid--deformable fusion strategies often involve complex operations such as interpolation, transformation composition, and inversion; such approaches are not included in our comparisons because of implementation complexity. In our method, a simple additive fusion strategy is adopted to combine rigid and deformable fields. While effective in practice, this approach does not explicitly constrain the interaction between the two components, which may theoretically lead to local inconsistencies. In addition, the current framework relies on a multi-stage pipeline, including segmentation and mask extraction, which increases system complexity. Future work will explore the use of higher-quality datasets to further validate model performance, as well as improved MRI data for accurate soft-tissue segmentation, enabling more comprehensive evaluation. Moreover, end-to-end joint learning frameworks that integrate segmentation and registration will be investigated to enhance clinical applicability.

In summary, we propose a rigid--deformable hybrid registration method, MSR. By fusing rigid deformation fields generated by the rigid module with deformable fields from the deformable branch, the model produces a unified hybrid transformation. The rigid module explicitly enforces constraints on vertebral structures from the perspective of deformation-field generation, while the deformable module, enhanced by MSL, effectively captures both global and local anatomical structures. As a result, MSR achieves accurate and robust cervical CT--MRI registration. In addition, we construct and publicly release the R-D-Reg dataset, providing a standardized and reproducible benchmark that facilitates fair comparison and further advances in this field.

\section*{CRediT authorship contribution statement}

\textbf{Bohai Zhang}: Conceptualization, Methodology, Software, Writing -- original draft, Investigation, Formal analysis, and Data curation.\par
\textbf{Wenjie Chen}: Visualization and Writing -- review \& editing.\par
\textbf{Mu Li}: Visualization.\par
\textbf{Kaixing Long}: Supervision and Writing -- review \& editing.\par
\textbf{Xing Shen}: Writing -- review \& editing.\par
\textbf{Xinqiang Yao}: Writing -- review \& editing.\par
\textbf{Jincheng Yang}: Writing -- review \& editing.\par
\textbf{Jianting Chen}: Writing -- review \& editing.\par
\textbf{Wei Yang}: Writing -- review \& editing.\par
\textbf{Qianjin Feng}: Writing -- review \& editing.\par
\textbf{Lei Cao}: Supervision, Project administration, Funding acquisition, Writing -- review \& editing.
\section*{Declaration of competing interest}

The authors declare that they have no known competing financial interests or personal relationships that could have appeared to influence the work reported in this paper.

\section*{Data availability}

I have shared the link of the involved data and code.

\section*{Acknowledgements}

This work was supported by the Guangdong Provincial Science and Technology Program
(Grant No.~2017B\allowbreak010110012) and the National College Students'
Innovation and Entrepreneurship Training Program (Grant No.~202512\allowbreak121061,
2025).

\bibliographystyle{cas-model2-names}

\bibliography{cas-refs}

\end{document}